\documentclass[runningheads]{llncs}
\usepackage[T1]{fontenc}
\usepackage{graphicx}
\usepackage{float}
\usepackage{multirow}

\begin{document}
\title{A Quantitative Evaluation Framework for Temporal Explainability in Echocardiographic Video Segmentation}

\author{Jiyoo Noh\inst{1} \and
Jonathan H. Chan\inst{2}}
\authorrunning{J. Noh et al.}
\institute{University of Toronto, Canada \and
King Mongkut's University of Technology Thonburi, Thailand
\email{jiyoo.noh@mail.utoronto.ca, jonathan@sit.kmutt.ac.th}
}
\maketitle              % typeset the header of the contribution
\begin{abstract}
Deep learning has achieved state-of-the-art performance in echocardiographic video segmentation, with an increasing number of models incorporating temporal information. However, quantitative evaluation of temporal explainability remains largely unexplored. We propose a quantitative framework for evaluating Grad-CAM explanations using four complementary metrics measuring temporal consistency, saliency motion, anatomical overlap, and temporal overlap. Using EchoNet-Dynamic, we compare a baseline 2D U-Net with ConvLSTM U-Net models trained across multiple temporal strides. While segmentation performance remained comparable across all models, intermediate ConvLSTM explanations exhibited substantially lower saliency consistency and greater centroid motion than final prediction explanations. Temporal Bottleneck explanations were significantly more stable than Encoder Bottleneck explanations across all strides, while final ConvLSTM Decoder3 explanations were broadly comparable to those of the 2D U-Net. Importantly, conventional frame-wise explanation metrics cannot determine whether variation in intermediate explanations reflects meaningful temporal feature evolution or explanation instability. These findings establish a preliminary quantitative framework for temporal explainability and motivate temporal-aware XAI methods that explicitly account for evolving representations in medical video models.

\end{abstract}

\section{Introduction}

Deep learning has become the dominant approach for medical image segmentation, achieving state-of-the-art performance across many imaging modalities \cite{ronneberger2015u,gipivskis2024explainable}. In echocardiography, convolutional neural networks accurately segment the left ventricle (LV) to enable automated estimation of clinically important indices such as ejection fraction \cite{ouyang2020video,thomas2022light}. More recently, temporal architectures such as ConvLSTMs \cite{shi2015convolutional} have been introduced to exploit motion information across consecutive frames.

Despite these advances, the interpretability of temporal segmentation models remains largely unexplored \cite{kolarik2023explainability,gipivskis2024explainable}. Explainable artificial intelligence (XAI) methods such as Gradient-weighted Class Activation Mapping (Grad-CAM) \cite{selvaraju2017grad} were originally developed for static images and are typically applied independently to each video frame \cite{kolarik2023explainability}. Consequently, existing evaluations rely primarily on qualitative visualization, providing limited insight into how explanations evolve throughout a cardiac cycle.

%In this work, we present a quantitative framework for evaluating temporal explainability in echocardiographic video segmentation. Using the EchoNet-Dynamic dataset \cite{Ouyang2020EchoNet}, we compare a conventional 2D U-Net \cite{Ronneberger2015} with ConvLSTM U-Net \cite{Shi2015ConvLSTM} models trained using multiple temporal sampling intervals. Grad-CAM explanations for each model are evaluated using four complementary temporal saliency metrics that measure explanation stability, spatial motion, anatomical localization, and temporal overlap. Rather than proposing a new explainability algorithm, our objective is to investigate whether temporal modeling naturally produces more temporally consistent explanations and to establish quantitative benchmarks for future temporal XAI research.

To address this gap, we propose a quantitative framework for evaluating temporal explainability in echocardiographic video segmentation. Unlike prior work that relies primarily on qualitative visualization, our framework provides a reproducible quantitative protocol for comparing temporal explanations across video segmentation models. We evaluate Grad-CAM explanations using four complementary temporal saliency metrics that measure explanation stability, spatial motion, anatomical localization, and temporal overlap.

To demonstrate the framework, we compare a conventional 2D U-Net \cite{ronneberger2015u} and a ConvLSTM U-Net \cite{shi2015convolutional} trained on the EchoNet-Dynamic dataset \cite{ouyang2020video} using multiple temporal sampling intervals. These models were intentionally selected as representative spatial and temporal baselines, as the primary contribution of this work is the evaluation of temporal explainability rather than the development of a new segmentation architecture.

The main contributions of this work are as follows:

\begin{itemize}
    \item We introduce a quantitative framework for evaluating temporal Grad-CAM explanations in echocardiographic video segmentation using complementary temporal saliency metrics.
    
    \item We compare representative spatial (2D U-Net) and temporal (ConvLSTM U-Net) segmentation models to investigate how temporal modeling influences explanation dynamics.

    \item We establish a preliminary evaluation protocol for temporal explainability in echocardiographic segmentation, providing a foundation for future temporal XAI methods.
\end{itemize}

\section{Related Work}

\subsection{Echocardiographic Video Segmentation}

%Deep learning has become the dominant approach for automated echocardiographic segmentation, with U-Net \cite{Ronneberger2015} serving as the standard architecture due to its encoder--decoder design and skip connections, which enable accurate localization of cardiac structures.

%The introduction of the EchoNet-Dynamic dataset \cite{Ouyang2020EchoNet}, containing over 10,000 annotated echocardiography videos with expert left ventricular tracings at the end-diastolic (ED) and end-systolic (ES) frames, established a standardized benchmark for developing and evaluating deep learning methods for cardiac video analysis. 

%Building upon these advances, recent methods have incorporated temporal information through recurrent, spatiotemporal, and attention-based architectures, such as ConvLSTM U-Net \cite{Shi2015ConvLSTM}, to improve segmentation robustness by leveraging cardiac motion across consecutive frames. %insert TAM, MemSAM citations
%However, existing work has largely focused on improving segmentation accuracy, while the effect of temporal modeling on model interpretability remains largely unexplored.

U-Net \cite{ronneberger2015u} has become a standard architecture for echocardiographic image segmentation due to its encoder--decoder design and skip connections, while the EchoNet-Dynamic dataset \cite{ouyang2020video} established a standardized benchmark for video-based cardiac segmentation using over 10,000 annotated echocardiography videos with expert left ventricular tracings at the end-diastolic (ED) and end-systolic (ES) frames. More recent approaches have incorporated temporal information through recurrent, spatiotemporal, and attention-based architectures to exploit cardiac motion across consecutive frames. ConvLSTM \cite{shi2015convolutional}, although not specific to echocardiography, extends the fully connected LSTM (FC-LSTM) by incorporating convolutional structures in both the input-to-state and state-to-state transitions. Bi-Directional ConvLSTM U-Net \cite{azad2019bi} builds upon this work by combining encoder and decoder feature maps rather than relying on simple concatenation in the skip connections of U-Net. More specifically to echocardiography, MV-RAN \cite{li2020mv} uses hierarchical ConvLSTM recurrent units to aggregate spatiotemporal features, while Lin et al. \cite{lin2024dynamic} use optical flow estimation (OFE) and dynamic-guided spatiotemporal attention (DSA) for semi-supervised echocardiography video segmentation. 

\subsection{Explainable AI for Medical Image Segmentation}

As deep learning models become increasingly prevalent in medical image segmentation, explainable artificial intelligence (XAI) has emerged as an important tool for improving model transparency and supporting clinical trust. Most XAI methods for medical image segmentation are post-hoc approaches that attribute predictions to salient image regions without modifying the underlying network \cite{gipivskis2024explainable}. Among these, Grad-CAM \cite{selvaraju2017grad}  remains one of the most widely used techniques, with extensions such as Grad-CAM++ \cite{chattopadhay2018grad} and Score-CAM \cite{wang2020score} improving localization accuracy. However, these methods are typically applied independently to each image or video frame, providing little insight into the temporal consistency of explanations for video segmentation models.

\subsection{Temporal Explainability in Medical Video Analysis}

Although XAI has been extensively studied for static medical images, comparatively little work has investigated explainability for video-based medical imaging. Specifically, temporal explainability remains comparatively underexplored in medical video analysis \cite{kolarik2023explainability}. Existing studies have primarily focused on cardiac function assessment rather than segmentation. For example, EchoGNN \cite{mokhtariechognn} estimates frame importance for ejection fraction prediction, while Er et al. \cite{er2025spatiotemporal} use a gradient-based saliency method to quantitatively evaluate video regression models' focus on the LV for ejection fraction (EF) prediction. Although these studies demonstrate the value of temporal explanations, they neither address video segmentation nor provide a quantitative framework for evaluating the temporal behavior of Grad-CAM explanations. Furthermore, current temporal XAI methods, largely visualize salient regions across individual frames rather than explicitly explaining how temporal dynamics contribute to model decisions. Consequently, it remains unclear whether existing frame-wise explanation methods adequately capture the temporal reasoning learned by video segmentation models. This gap motivates the framework proposed in this work.

\section{Methods}

\begin{figure}[H]
    \centering
    \includegraphics[width=\textwidth]{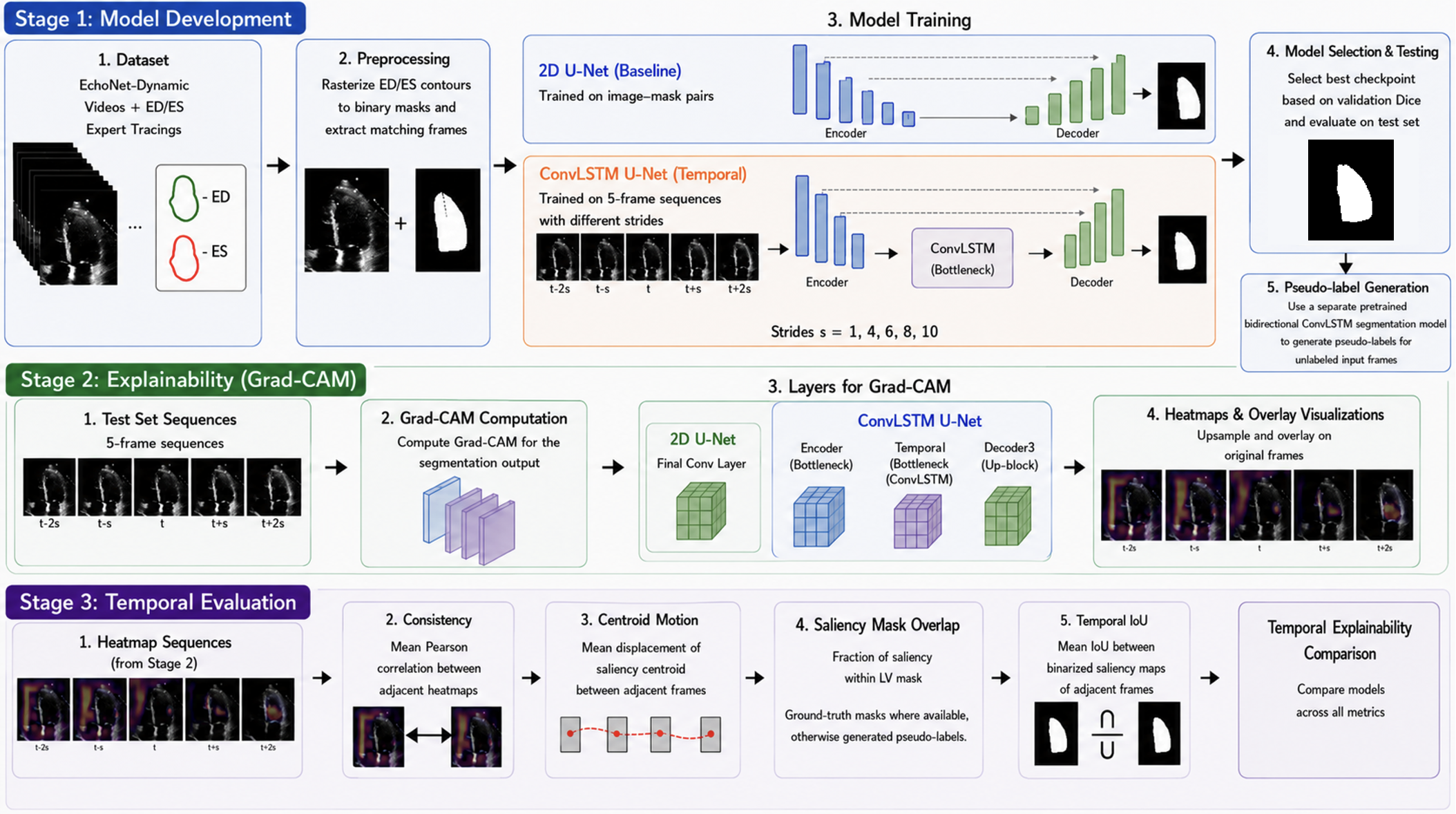}
    \caption{Overview of the proposed workflow, including data preprocessing, model training, Grad-CAM generation, and quantitative temporal evaluation of explanation consistency and anatomical relevance.}
    \label{fig:full_pipeline}
\end{figure}

\subsection{Framework Overview}

Figure~\ref{fig:full_pipeline} illustrates the proposed framework. The EchoNet-Dynamic dataset \cite{ouyang2020video} provides expert left ventricular (LV) contours at the end-diastolic (ED) and end-systolic (ES) frames of each video. These contours are rasterized into binary segmentation masks and paired with the corresponding grayscale images to train a baseline 2D U-Net \cite{ronneberger2015u}. The original videos are retained for training ConvLSTM U-Net models using five-frame temporal sequences.

Since EchoNet-Dynamic provides manual annotations only for the ED and ES frames, pseudo-labels are generated for all unlabeled frames using a separately trained bidirectional ConvLSTM segmentation model. Ground-truth masks are retained for the annotated ED/ES frames, while the generated pseudo-labels are used for the remaining frames during explainability evaluation, enabling frame-specific anatomical supervision across the entire input sequence.

The baseline U-Net predicts the LV segmentation mask from a single frame using a combined Dice and Binary Cross-Entropy (BCE) loss,

\begin{equation}
\mathcal{L}
=
\mathcal{L}_{\rm Dice}
+
\mathcal{L}_{\rm BCE}.
\end{equation}

To incorporate temporal information, the ConvLSTM U-Net replaces the U-Net bottleneck with a ConvLSTM module that aggregates encoder features across a five-frame sequence,

\begin{equation}
X=
\{I_{t-2s},I_{t-s},I_t,I_{t+s},I_{t+2s}\},
\end{equation}

where $s$ denotes the temporal stride. Each frame is encoded using a shared U-Net encoder, after which the ConvLSTM aggregates the bottleneck features and the final hidden state is passed to the decoder together with the skip connections from the center frame. Models are trained using strides of $s\in\{1,4,6,8,10\}$ to investigate the influence of temporal context.

To visualize the spatial regions influencing segmentation predictions, Grad-CAM \cite{selvaraju2017grad} is applied to the trained models to generate explanation heatmaps. For the baseline 2D U-Net, Grad-CAM is computed independently for each frame using the corresponding frame-specific ground-truth or pseudo-label mask as the target, and the resulting heatmaps are stacked to form a temporal explanation sequence.

For the ConvLSTM U-Net, Grad-CAM is generated from the encoder bottleneck, temporal bottleneck, and Decoder3, enabling visualization of feature representations before temporal aggregation, after temporal fusion, and during decoder reconstruction. Because the encoder bottleneck and temporal bottleneck are evaluated once for each frame in the input sequence, Grad-CAM produces a unique heatmap for every temporal position. In contrast, Decoder3 is evaluated once after temporal aggregation and therefore produces a single heatmap per inference.

To enable quantitative temporal evaluation of Decoder3, the inference procedure is repeated for each frame in the sequence, with the target frame shifted at each iteration and the corresponding five-frame input sequence reconstructed around the target frame. The resulting frame-specific Decoder3 heatmaps are then stacked to form a temporal explanation sequence. As shown in Figure~\ref{fig:full_pipeline}, the resulting heatmap sequences serve as the input for the quantitative temporal explainability evaluation described in Section~3.2.

\subsection{Temporal Explanation Evaluation}

To quantitatively assess the temporal behavior of Grad-CAM explanations, four temporal saliency metrics are computed for each generated heatmap sequence.

\textbf{Saliency Consistency} measures the similarity between consecutive Grad-CAM heatmaps using the Pearson correlation coefficient,

\begin{equation}
C = \frac{1}{T-1}\sum_{t=1}^{T-1}
\mathrm{corr}(H_t,H_{t+1}),
\end{equation}

where $H_t$ denotes the normalized Grad-CAM heatmap at frame $t$, and $T=5$ is the sequence length. Higher values indicate greater temporal stability of the explanations.

\textbf{Saliency Centroid Motion} quantifies the spatial displacement of the center of attention between consecutive frames. The centroid of each heatmap is computed as the saliency-weighted center of mass,

\begin{equation}
\mathbf{c}_t=
\frac{\sum_{i,j}(i,j)H_t(i,j)}
{\sum_{i,j}H_t(i,j)},
\end{equation}

and the average centroid displacement is calculated as

\begin{equation}
M=\frac{1}{T-1}\sum_{t=1}^{T-1}
\|\mathbf{c}_{t+1}-\mathbf{c}_t\|_2.
\end{equation}

%\textbf{Center Saliency Mask Overlap} measures the proportion of Grad-CAM activation contained within the ground-truth left ventricular segmentation mask. This metric is computed using the ground truth segmentation mask for the labelled ED/ES frames, and for the unlabelled frames, using the generated pseudolabels.

\textbf{Saliency Mask Overlap} measures the proportion of Grad-CAM activation contained within the left ventricular segmentation mask. Ground-truth segmentation masks are used where available, while pseudo-labels are used for unannotated frames. Although saliency mask overlap may partly reflect the nature of the segmentation task or network architecture, it is included as one of four complementary evaluation metrics to characterize different aspects of temporal explanation behaviour, rather than as a standalone measure of explanation faithfulness.
%Because EchoNet-Dynamic provides annotations only for the ED/ES frame, this metric is computed exclusively for the center frame.
%\begin{equation}
%O=
%\frac{\sum_{(i,j)\in\Omega}H_c(i,j)}
%{\sum_{i,j}H_c(i,j)},
%\end{equation}

%where $\Omega$ denotes the ground-truth left ventricular mask and $H_c$ is the Grad-CAM heatmap of the center frame.

Finally, \textbf{Temporal Saliency IoU} measures the overlap between thresholded saliency regions in consecutive frames using the standard Intersection-over-Union (IoU) metric, where the saliency region is defined as the top 20\% of normalized Grad-CAM values.

Together, these metrics quantify the temporal stability and anatomical consistency of Grad-CAM explanations.

\section{Experiments}

\subsection{Dataset and Experimental Setup}

Experiments were conducted on the EchoNet-Dynamic dataset \cite{ouyang2020video} using the official training, validation, and test split. All models were optimized with AdamW using the Dice/BCE loss described in Section~3.1. ConvLSTM models employed early stopping with best-checkpoint selection based on validation Dice. The baseline was evaluated using test loss and Dice, while ConvLSTM models were additionally evaluated using IoU, precision, and recall.

\subsection{Compared Models}

%Six segmentation models were evaluated in this study. As a spatial baseline, a conventional 2D U-Net was trained using individual echocardiography frames. Temporal segmentation models consisted of a ConvLSTM U-Net trained using five-frame input sequences centered on the annotated ED or ES frame. A baseline temporal model was first trained using consecutive frames ($s=1$), followed by four additional models with temporal strides of $s\in\{4,6,8,10\}$. All ConvLSTM models shared the same network architecture and training procedure, differing only in the temporal spacing between adjacent frames within the input sequence. This design enabled a controlled evaluation of how increasing temporal context influences both segmentation performance and the temporal consistency of Grad-CAM explanations. 

Six segmentation models were evaluated in this study. As a spatial baseline, a baseline 2D U-Net was trained on individual frames, and as a temporal baseline, five ConvLSTM U-Net models were trained using temporal strides of $s\in\{1,4,6,8,10\}$. All ConvLSTM models shared the same architecture and training procedure described in Section~3.1, differing only in temporal stride. This design enabled an isolated evaluation of the effect of temporal context on segmentation performance and explainability.

\subsection{Explainability Evaluation Protocol}

For each model, the checkpoint achieving the highest validation Dice score was selected for explainability analysis. Since EchoNet-Dynamic provides manual annotations only for the ED and ES frames, pseudo-labels generated by a pretrained bidirectional ConvLSTM U-Net were used for all unannotated frames. This model was selected as it achieved high segmentation performance on EchoNet-Dynamic and provided anatomically plausible frame-level masks for unannotated frames. Ground-truth masks were retained where available.

The Grad-CAM generation procedure followed the framework described in Section~3.1. Heatmap sequences were generated for the baseline 2D U-Net, encoder bottleneck, temporal bottleneck, and Decoder3. All normalized heatmap sequences were saved as NumPy arrays together with visualization overlays and evaluated using the temporal explainability metrics described in Section~3.2.

%For each trained model, the checkpoint achieving the highest validation Dice score was selected for explainability analysis. Grad-CAM was generated for every sample in the held-out test set using the corresponding best model checkpoint. For the baseline 2D U-Net, Grad-CAM was computed independently for each frame in the five-frame temporal sequence using the network's final convolutional layer, after which the resulting heatmaps were stacked to form a temporal explanation sequence.

%For each ConvLSTM U-Net model, Grad-CAM was evaluated at three representative network locations: the encoder bottleneck (\texttt{bottleneck\_encoder}), the temporal ConvLSTM bottleneck (\texttt{temporal\_bottleneck}), and the third decoder block (\texttt{decoder3}). These layers were selected to analyze the evolution of feature representations before temporal aggregation, after temporal feature fusion, and during decoder reconstruction.

%For every test sample, the normalized Grad-CAM heatmaps were saved as both NumPy arrays and individual heatmap images. In addition, overlay visualizations were generated by superimposing each heatmap onto its corresponding ultrasound frame, producing a temporal visualization of model attention across the input sequence. The resulting heatmaps served as the input for the quantitative temporal explanation evaluation described in Section~3.5.

\section{Results}
\subsection{Segmentation Performance}

Table \ref{tab:segmentation_results}  summarizes segmentation performance on the EchoNet-Dynamic test set. All models achieved comparable performance, with Dice scores above 0.91 and IoU values around 0.85. The ConvLSTM models slightly outperformed the baseline 2D U-Net, although differences between temporal stride configurations were negligible. The stride-8 model achieved the highest Dice (0.917) and IoU (0.849). Overall, segmentation accuracy remained largely unchanged across models, allowing subsequent differences in explainability to be interpreted independently of segmentation performance.

%Table~\ref{tab:segmentation_results} summarizes the segmentation performance of the baseline 2D U-Net and the ConvLSTM U-Net models across different temporal strides. All models achieved strong segmentation accuracy on the EchoNet-Dynamic test set, with Dice scores above 0.91 and IoU values around 0.85. The baseline 2D U-Net achieved a test Dice score of 0.9102, while all ConvLSTM variants achieved slightly higher Dice scores ranging from 0.9159 to 0.9167. The stride-8 model achieved the highest overall performance, with a test Dice of 0.9167, test IoU of 0.8485, and the lowest test loss of 0.1021. However, the performance differences among the ConvLSTM variants were small, indicating that increasing the temporal stride had little effect on conventional segmentation accuracy. The consistently high performance across all models provides a strong foundation for the subsequent explainability analysis, allowing differences in Grad-CAM localization and temporal saliency stability to be examined independently of segmentation quality.

\begin{table}[t]
\centering
\caption{Segmentation performance on the EchoNet-Dynamic test set.}
\label{tab:segmentation_results}
\begin{tabular}{lccc}
\hline
Model & Test Dice & Test IoU & Test Loss \\
\hline
2D U-Net & 0.910& N/A& 0.111\\
ConvLSTM (Stride 1) & 0.916& 0.847& 0.103\\
ConvLSTM (Stride 4) & 0.917& 0.848& 0.102\\
ConvLSTM (Stride 6) & 0.916& 0.848& 0.103\\
ConvLSTM (Stride 8) & \textbf{0.917}& \textbf{0.849}& \textbf{0.102}\\
ConvLSTM (Stride 10) & 0.917& 0.848& 0.103\\
\hline
\end{tabular}
\end{table}
\subsection{Qualitative Final Prediction Explanation Analysis}

%Representative Grad-CAM visualizations for the baseline 2D U-Net and ConvLSTM U-Net models are shown in Figure \ref{fig:gradcam_stride_comparison}. Across all models, the highest Grad-CAM activations were concentrated within the left ventricular (LV) cavity, indicating that the networks based their segmentation decisions on anatomically relevant image regions.

%Representative Grad-CAM visualizations for Decoder3 and the correspinding predicted masks are shown in Figure \ref{fig:Decoder_3_comparison_updated}. Across all models, the highest activations were concentrated within the left ventricular cavity, indicating anatomically relevant explanations. Compared with the baseline U-Net, ConvLSTM models produced more localized saliency maps that more closely followed the LV cavity, while differences between temporal stride configurations were minimal. These qualitative observations suggest that incorporating temporal information improves the localization of model explanations while preserving segmentation performance.

\begin{figure}[H]
    \centering
    \includegraphics[width=\textwidth]{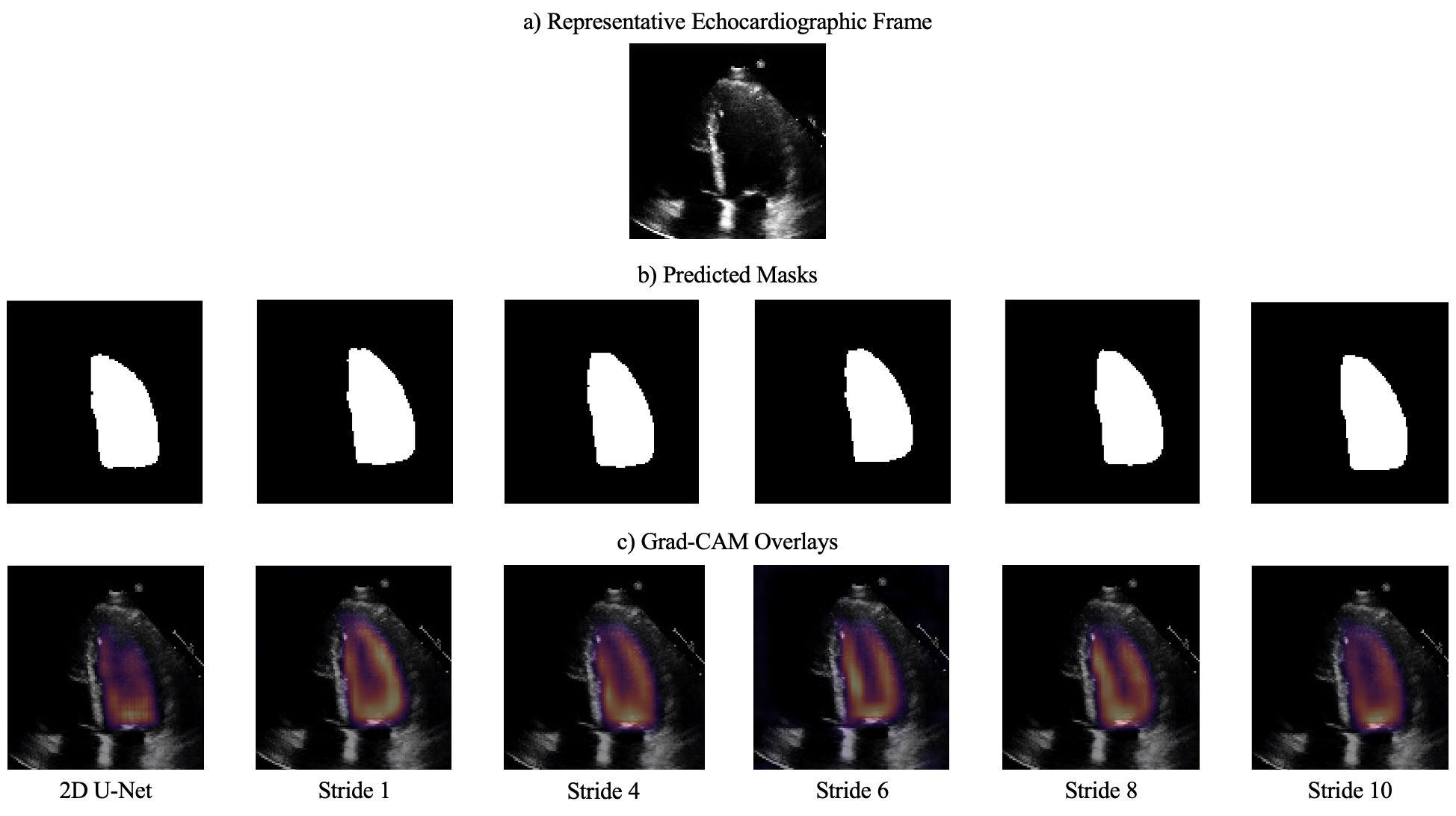}
    \caption{Qualitative comparison of final prediction explanations for the baseline 2D U-Net and ConvLSTM U-Net models. All visualizations correspond to the same patient and target frame and all ConvLSTM Grad-CAM heatmaps are from the Decoder3 layer.}
    \label{fig:Decoder_3_comparison_updated}
\end{figure}

Representative final prediction explanations are shown in Figure \ref{fig:Decoder_3_comparison_updated}. For each model, the predicted LV segmentation mask and corresponding Grad-CAM overlay are presented for the same target frame, enabling a direct comparison of segmentation predictions and the final spatial explanations produced by each network. 

Across all models, the predicted segmentation masks closely align with the LV cavity, consistent with the comparable segmentation performance reported in Section 5.1. Similarly, the highest activations in the Grad-CAM overlays are concentrated primarily within the LV cavity, indicating that both the baseline U-Net and ConvLSTM models base their final predictions on anatomically relevant regions. Compared with the baseline 2D U-Net, the ConvLSTM Decoder3 explanations appear slightly more localized around the LV cavity, although qualitative differences between temporal stride configurations are minimal. 

These results suggest that incorporating temporal information preserves anatomically meaningful final prediction explanations while producing only modest visual differences across temporal strides. Because the final explanations are qualitatively similar, quantitative analysis is required to determine whether these differences are statistically significant and to assess how temporal reasoning influences the internal feature representations examined in the following sections.

\subsection{Qualitative Analysis of Internal Representation Evolution}

\begin{figure}[H]
    \centering
    \includegraphics[width=0.95\textwidth]{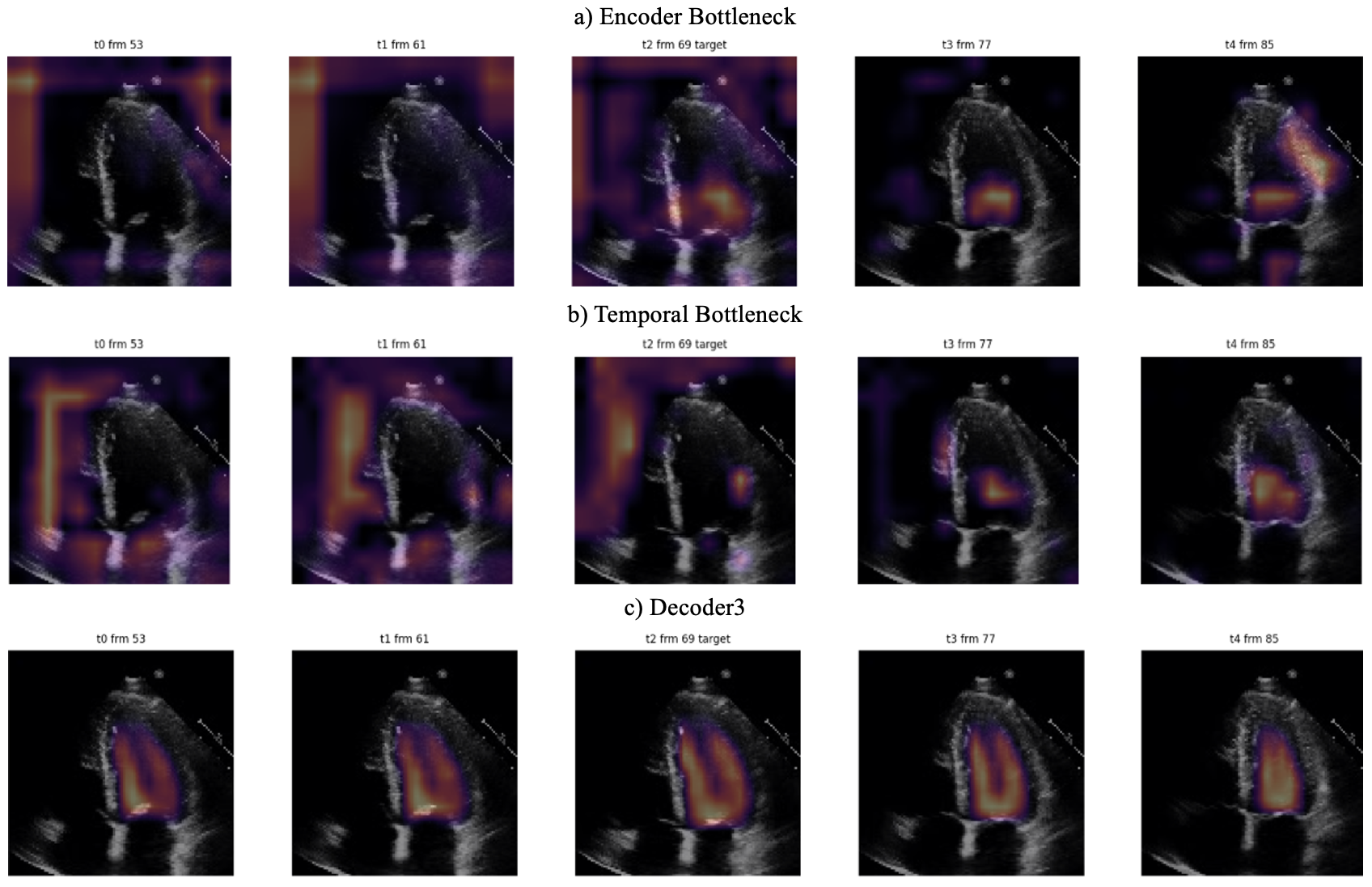}
    \caption{Evolution of Grad-CAM explanations across the ConvLSTM Stride 8 network. Representative Grad-CAM heatmaps from the encoder bottleneck, temporal bottleneck, and Decoder3 are shown for all five frames of the input sequence.}
    \label{fig:gradcam_layer_comparison}
\end{figure}

While Figure \ref{fig:Decoder_3_comparison_updated} compares the final prediction explanations across models, Figure \ref{fig:gradcam_layer_comparison} illustrates how these explanations are formed within the ConvLSTM network. Grad-CAM visualizations from the Encoder Bottleneck, Temporal Bottleneck, and Decoder3 layers are shown across the five-frame input sequence for the ConvLSTM Stride 8 model. The Stride 8 model is shown for the representative visualization due to its superior performance among the ConvLSTM models.

The Encoder Bottleneck produces coarse and spatially diffuse activations that vary substantially between frames. After temporal aggregation, the Temporal Bottleneck generates more localized responses while continuing to evolve throughout the sequence. In contrast, Decoder3 produces the most anatomically focused explanations, consistently aligning with the LV cavity across all five frames.

These qualitative observations suggest that ConvLSTM representations become progressively more localized throughout the network, ultimately producing anatomically meaningful final prediction explanations. The following section quantitatively evaluates these differences using the proposed temporal explainability metrics.

\subsection{Quantitative Temporal Explainability Evaluation}

%To quantitatively evaluate the temporal behavior of Grad-CAM explanations, four complementary metrics were computed: saliency consistency, saliency centroid motion, center saliency mask overlap, and temporal saliency IoU. These metrics were evaluated using the bottleneck encoder and temporal bottleneck layers of the ConvLSTM models, as both layers produce frame-specific feature representations throughout the input sequence. The decoder layer was excluded from the quantitative temporal evaluation because its Grad-CAM explanations are generated after temporal feature fusion, producing a single fused representation rather than independent frame-specific activations. Consequently, frame-to-frame temporal metrics are not directly applicable to this layer.

%Table~\ref{tab:temporal_metrics} summarizes the temporal explainability metrics for the baseline 2D U-Net and all ConvLSTM models. Across the evaluated temporal strides, the ConvLSTM models exhibited broadly similar temporal behavior, indicating that increasing the temporal sampling interval had only a modest influence on explanation dynamics. No single stride consistently outperformed the others across all four metrics, suggesting that temporal stride has relatively little effect on the stability of the learned explanations.

%Table \ref{tab:temporal_metrics} summarizes the quantitative temporal explainability results. Metrics were computed for the encoder and temporal bottleneck layers, while Decoder3 was excluded because its temporally fused representations are not directly comparable across individual frames.

Table \ref{tab:internal_representations} summarizes the quantitative evaluation of the ConvLSTM Bottleneck Encoder and Temporal Bottleneck representations across all temporal strides. Consistent with the qualitative observations in Figure \ref{fig:gradcam_layer_comparison}, both intermediate representations exhibited substantially lower saliency consistency and temporal IoU, in addition to higher centroid motion, when compared with the final prediction explanations presented in Section 5.2. These findings indicate greater temporal variation in the explanations generated from the internal feature representations. However, the current analysis cannot distinguish whether this variation reflects meaningful temporal feature evolution or instability in the Grad-CAM explanations.

\begin{table}[t]
\centering
\caption{Quantitative evaluation of ConvLSTM intermediate representations across temporal strides. Results are reported for the Encoder Bottleneck and Temporal Bottleneck layers. }
\label{tab:internal_representations}
\resizebox{\linewidth}{!}{
\begin{tabular}{cccccc}
\hline
\textbf{Stride} &
\textbf{Layer} &
\textbf{Consistency $\uparrow$} &
\textbf{Motion $\downarrow$} &
\textbf{Overlap $\uparrow$} &
\textbf{IoU $\uparrow$} \\
\hline

\multirow{2}{*}{1}
& Encoder Bottleneck
& 0.179
& 19.942
& 0.112
& 0.278 \\

& Temporal Bottleneck
& \textbf{0.309}
& \textbf{11.100}
& 0.067
& 0.300 \\

\hline

\multirow{2}{*}{4}
& Encoder Bottleneck
& 0.087
& 20.574
& 0.164
& 0.232 \\

& Temporal Bottleneck
& \textbf{0.192}
& \textbf{11.868}
& 0.044
& 0.212 \\

\hline

\multirow{2}{*}{6}
& Encoder Bottleneck
& 0.112
& 22.834
& 0.166
& 0.196 \\

& Temporal Bottleneck
& \textbf{0.186}
& \textbf{13.553}
& \textbf{0.195}
& 0.177 \\

\hline

\multirow{2}{*}{8}
& Encoder Bottleneck
& 0.123
& 20.048
& \textbf{0.278}
& \textbf{0.215} \\

& Temporal Bottleneck
& \textbf{0.238}
& \textbf{15.141}
& 0.165
& 0.182 \\

\hline

\multirow{2}{*}{10}
& Encoder Bottleneck
& 0.105
& 21.823
& \textbf{0.291}
& \textbf{0.279} \\

& Temporal Bottleneck
& \textbf{0.267}
& \textbf{11.816}
& 0.134
& 0.206 \\

\hline
\end{tabular}
}

\vspace{0.5em}
\footnotesize
Paired Wilcoxon signed-rank tests with Holm--Bonferroni correction demonstrated that the Temporal Bottleneck achieved significantly higher saliency consistency and significantly lower saliency centroid motion than the Encoder Bottleneck across all temporal strides (all adjusted $p<0.001$). Statistical significance for saliency mask overlap and temporal saliency IoU varied across temporal strides.
\end{table}

Across all temporal strides, the Temporal Bottleneck showed higher saliency consistency and lower centroid motion than the Encoder Bottleneck, indicating more temporally stable explanations at this layer. However, both internal layers remained substantially less consistent than the final prediction explanations. Differences between temporal stride configurations were modest, suggesting that these trends were largely preserved across sampling intervals.
\begin{figure}[!t]
    \centering
    \includegraphics[width=0.95\textwidth]{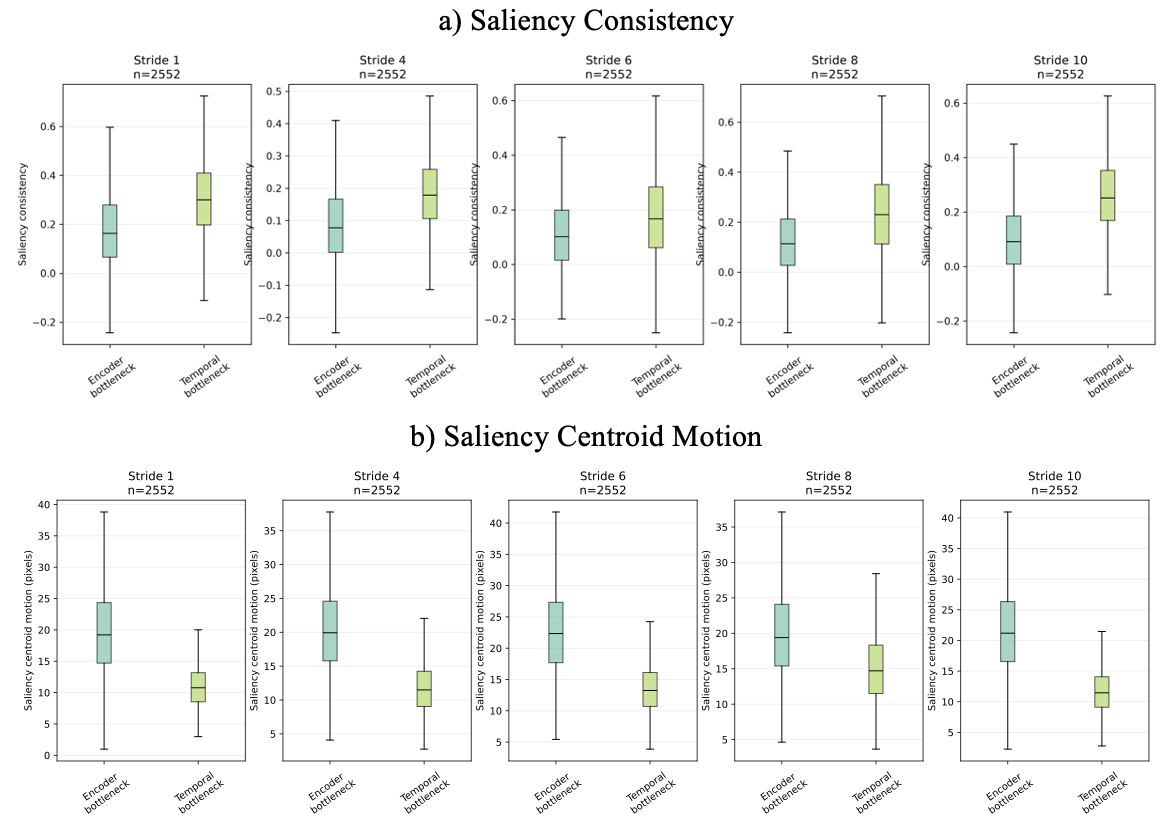}
    \caption{Distribution of saliency consistency (top) and saliency centroid motion (bottom) for the Encoder Bottleneck and Temporal Bottleneck across all temporal strides. Boxes represent the interquartile range, the center line indicates the median, and whiskers extend to the most extreme non-outlier observations.}
    \label{fig:internal_boxplots}
\end{figure}

Figure~\ref{fig:internal_boxplots} further illustrates the distribution of saliency consistency and saliency centroid motion across all 2,552 test samples. Compared with the Encoder Bottleneck, the Temporal Bottleneck consistently shifts toward higher saliency consistency and lower centroid motion for every temporal stride, reinforcing the statistical trends reported in Table~\ref{tab:internal_representations}. Nevertheless, these intermediate representations remain substantially less stable than the final prediction explanations.

Taken together, the qualitative observations in Figure~\ref{fig:gradcam_layer_comparison} and the quantitative results presented here demonstrate that conventional frame-wise Grad-CAM metrics report substantially lower temporal consistency for intermediate ConvLSTM representations than for the final prediction explanations. However, these metrics cannot distinguish whether this behaviour reflects the expected evolution of intermediate representations during temporal aggregation or genuine explanation instability. This ambiguity highlights an important limitation of applying frame-wise explainability metrics to temporal architectures and motivates the development of temporal-aware XAI methods capable of explicitly accounting for evolving feature representations.

\subsection{Quantitative Evaluation of Final Prediction Explanations}

Table~\ref{tab:final_prediction_explanations} summarizes the quantitative evaluation of the final prediction explanations produced by the baseline 2D U-Net and the ConvLSTM Decoder3 layer across all temporal strides. Unlike the intermediate representations presented in Section~5.4, the final prediction explanations achieved substantially higher saliency consistency, saliency mask overlap, and temporal IoU, indicating that both models produced anatomically meaningful and temporally coherent explanations.

\begin{table}[t]
\centering
\caption{Quantitative evaluation of final prediction explanations across temporal strides. Results are reported for the baseline 2D U-Net final convolution layer and the ConvLSTM Decoder3 layer.}
\label{tab:final_prediction_explanations}
\resizebox{\linewidth}{!}{
\begin{tabular}{cccccc}
\hline
\textbf{Stride} &
\textbf{Model} &
\textbf{Consistency $\uparrow$} &
\textbf{Motion $\downarrow$} &
\textbf{Overlap $\uparrow$} &
\textbf{IoU $\uparrow$} \\
\hline

\multirow{2}{*}{1}
& 2D U-Net
& 0.991
& 0.608
& \textbf{0.961}
& \textbf{0.996} \\

& ConvLSTM Decoder3
& \textbf{0.997}
& \textbf{0.438}
& 0.896
& 0.906 \\

\hline

\multirow{2}{*}{4}
& 2D U-Net
& 0.950
& 1.526
& \textbf{0.961}
& \textbf{0.991} \\

& ConvLSTM Decoder3
& \textbf{0.973}
& \textbf{0.996}
& 0.905
& 0.897 \\

\hline

\multirow{2}{*}{6}
& 2D U-Net
& 0.922
& 1.893
& \textbf{0.961}
& \textbf{0.990} \\

& ConvLSTM Decoder3
& \textbf{0.941}
& \textbf{1.646}
& 0.683
& 0.734 \\

\hline

\multirow{2}{*}{8}
& 2D U-Net
& 0.904
& 2.111
& \textbf{0.961}
& \textbf{0.988} \\

& ConvLSTM Decoder3
& \textbf{0.930}
& \textbf{1.500}
& 0.914
& 0.848 \\

\hline

\multirow{2}{*}{10}
& 2D U-Net
& 0.890
& 2.298
& \textbf{0.961}
& \textbf{0.987} \\

& ConvLSTM Decoder3
& \textbf{0.918}
& \textbf{1.708}
& 0.910
& 0.791 \\

\hline
\end{tabular}
}

\vspace{0.5em}

\footnotesize
Paired Wilcoxon signed-rank tests with Holm--Bonferroni correction identified statistically significant differences between the baseline 2D U-Net and ConvLSTM Decoder3 for all evaluated metrics across all temporal strides (all adjusted $p<0.001$). Although statistically significant, the absolute differences in saliency consistency and saliency centroid motion were generally modest, while larger differences were observed for saliency mask overlap and temporal saliency IoU.
\end{table}

Across all temporal strides, the ConvLSTM Decoder3 produced results broadly comparable to the baseline 2D U-Net, with only modest variation across stride configurations. Differences in saliency consistency and centroid motion were relatively small, whereas larger differences were observed for saliency mask overlap and temporal saliency IoU. Overall, the final prediction explanations were substantially more temporally stable than the intermediate-layer explanations.

\begin{figure}[!t]
    \centering
    \includegraphics[width=0.95\textwidth]{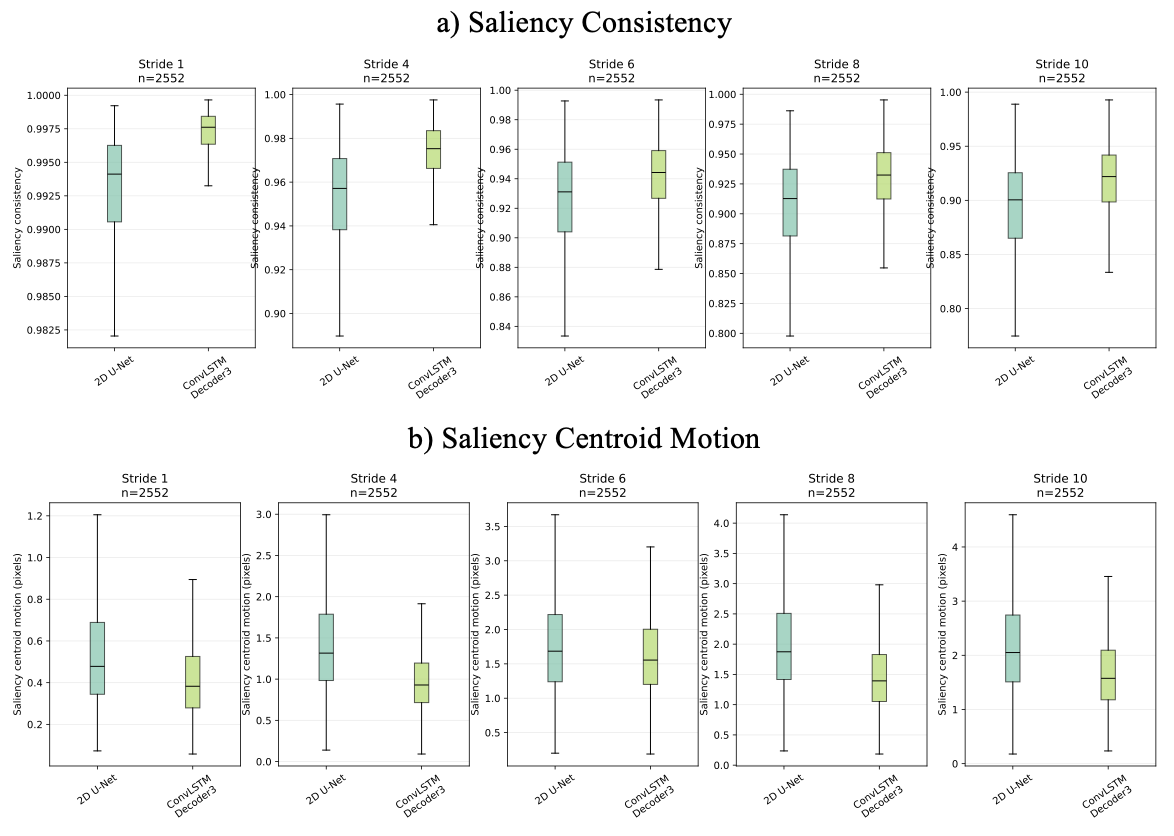}
    \caption{Distribution of saliency consistency (top) and saliency centroid motion (bottom) for the baseline 2D U-Net and ConvLSTM Decoder3 final prediction explanations across all temporal strides. Boxes represent the interquartile range, the center line indicates the median, and whiskers denote the full data range excluding outliers.}
    \label{fig:final_boxplots}
\end{figure}

Figure~\ref{fig:final_boxplots} illustrates the distribution of saliency consistency and saliency centroid motion across all 2,552 test samples. Compared with the intermediate representations shown in Figure~\ref{fig:internal_boxplots}, the distributions of the final prediction explanations are substantially more concentrated and exhibit only minor differences between the baseline 2D U-Net and ConvLSTM Decoder3 across temporal strides.

Together, the results from Sections~5.4 and~5.5 indicate that conventional frame-wise Grad-CAM metrics produce markedly different conclusions depending on which network representation is evaluated. While intermediate ConvLSTM representations appear substantially less temporally consistent than their final prediction explanations, the current evaluation framework cannot determine whether this behaviour reflects genuine explanation instability or the expected evolution of temporally aggregated feature representations. This limitation motivates the development of temporal-aware XAI methods capable of explicitly distinguishing evolving temporal reasoning from unstable explanations.

\section{Discussion}
This study introduced a quantitative framework for evaluating temporal explainability in echocardiographic video segmentation using Grad-CAM and four complementary temporal saliency metrics. While the intermediate ConvLSTM representations exhibited substantially lower temporal consistency and higher centroid motion than the baseline 2D U-Net, the final prediction explanations produced by the ConvLSTM Decoder3 layer remained highly temporally consistent and anatomically localized across all evaluated temporal strides.

Rather than indicating inferior explainability, these findings highlight a fundamental ambiguity when applying conventional frame-wise XAI evaluation to temporal architectures. Intermediate ConvLSTM representations are explicitly designed to evolve as temporal information is aggregated, whereas final prediction explanations are generated after temporal feature fusion. Conventional frame-wise Grad-CAM metrics cannot distinguish whether reduced temporal consistency in intermediate representations reflects genuine explanation instability or the expected evolution of temporally aggregated features. 

This finding has broader implications for medical video analysis. The interpretation of frame-wise explainability metrics depends strongly on which network representation is evaluated, and static evaluation metrics may therefore lead to different conclusions for intermediate and final explanations within the same model. These results motivate explainability methods and evaluation frameworks that explicitly account for evolving temporal representations rather than interpreting all temporal variation as explanation inconsistency. The proposed framework therefore provides a preliminary evaluation protocol for future temporal XAI methods, enabling both qualitative and quantitative comparison of temporal explanations.

Several limitations should be considered when interpreting these results. First, although the proposed framework quantitatively evaluates both intermediate and final prediction explanations, it cannot determine whether changes in intermediate saliency maps arise from expected temporal feature evolution or genuine explanation instability. Second, the evaluation of non-annotated frames relies on pseudo-labels generated by a pretrained segmentation model rather than expert manual annotations, which may introduce additional uncertainty into the quantitative metrics. Finally, the proposed framework was evaluated only using Grad-CAM on ConvLSTM U-Net models with fixed five-frame input sequences. Future work should extend this framework to additional explainability methods, temporal architectures, longer temporal contexts, and datasets with dense frame-level annotations.

\section{Conclusion and Future Work}

This work presented a quantitative framework for evaluating temporal explainability in echocardiographic video segmentation using Grad-CAM and four complementary temporal saliency metrics. Although segmentation performance was comparable across all models, the proposed framework revealed that conventional frame-wise explainability metrics produce markedly different conclusions depending on whether intermediate representations or final prediction explanations are evaluated. While intermediate ConvLSTM representations exhibited substantially lower temporal consistency than the final prediction explanations, the current evaluation framework cannot determine whether this behaviour reflects expected temporal feature evolution or genuine explanation instability. These findings highlight an important limitation of static explainability evaluation for temporal architectures and demonstrate the need for temporal-aware explainability methods and evaluation metrics that explicitly account for evolving feature representations in medical video models.

Future work should focus on developing temporal-aware explainability methods capable of explicitly modelling evolving feature representations in medical video models. Extending the proposed evaluation framework to additional explainability techniques, temporal network architectures, and larger multi-center datasets will help determine whether the observed trends generalize beyond ConvLSTM models and Grad-CAM. An important direction is to compare temporal explanations with independent measures of cardiac motion, such as optical flow or learned motion representations, to determine whether changes in intermediate saliency maps correspond to physiologically meaningful temporal reasoning or genuine explanation instability. Such methods could provide a more faithful characterization of temporal decision-making in deep learning models for echocardiographic video analysis.

%Future work should focus on developing temporal explainability methods that model evolving feature representations, while extending the proposed evaluation framework to additional XAI techniques, video segmentation architectures, and datasets to further improve the interpretability of temporal deep learning models.

%\section{Future Work}

%Future work should focus on developing a dedicated temporal XAI module for video segmentation models. Rather than applying frame-wise explanation methods independently to each frame, such a module should explicitly model how salient anatomical regions evolve across time and how temporal information contributes to the final segmentation prediction. This could involve temporally-aware saliency propagation, attention-based explanation tracking, or motion-guided attribution mechanisms that directly account for cardiac dynamics.

%In addition, future studies should evaluate the proposed framework using other XAI methods, longer temporal windows, and more advanced temporal architectures such as transformer-based video segmentation models. Extending the analysis to datasets with dense frame-level annotations would also enable more complete evaluation of anatomical localization throughout the cardiac cycle.


\bibliographystyle{splncs04}
\bibliography{references}

\end{document}